\documentclass{article}

    \PassOptionsToPackage{numbers, compress}{natbib}

    \usepackage[preprint]{neurips_2020}

\usepackage[utf8]{inputenc} 
\usepackage[T1]{fontenc}    
\usepackage{hyperref}       
\usepackage{url}            
\usepackage{booktabs}       
\usepackage{amsfonts}       
\usepackage{nicefrac}       
\usepackage{microtype}      

\usepackage{times}
\usepackage{epsfig}
\usepackage{graphicx}
\usepackage{amsmath}
\usepackage{amssymb}
\usepackage{authblk}
\usepackage{booktabs}

\usepackage{multirow}
\usepackage{caption}
\usepackage{array}
\usepackage{subfigure}
\title{Analogical Reasoning for Visually Grounded Language Acquisition}

 \author{
 \textbf{Bo Wu},
 \textbf{Haoyu Qin}, 
 \textbf{Alireza Zareian},
 \textbf{Carl Vondrick}, 
 \textbf{Shih-Fu Chang} \\
  Columbia University, New York NY 10027, USA\\
    bobbywu.cs@gmail.com, \{bo.wu, hq2172, az2407, sc250\}@columbia.edu, vondrick@cs.columbia.edu \\
  }

\begin{document}

\maketitle

\begin{abstract}
Children acquire language subconsciously by observing the surrounding world and listening to descriptions. They can discover the meaning of words even without explicit language knowledge, and generalize to novel compositions effortlessly. In this paper, we bring this ability to AI, by studying the task of Visually grounded Language Acquisition (VLA). 
We propose a multimodal transformer model augmented with a novel mechanism for analogical reasoning, which approximates novel compositions by learning semantic mapping and reasoning operations from previously seen compositions.  
Our proposed method, Analogical Reasoning Transformer Networks (\textsc{ARTNet}), is trained on raw multimedia data (video frames and transcripts), and after observing a set of compositions such as ``washing apple'' or ``cutting carrot'', it can generalize and recognize new compositions in new video frames, such as ``washing carrot'' or ``cutting apple''.
To this end, \textsc{ARTNet} refers to relevant instances in the training data and uses their visual features and captions to establish analogies with the query image. Then it chooses the suitable verb and noun to create a new composition that describes the new image best. Extensive experiments on an instructional video dataset demonstrate that the proposed method achieves significantly better generalization capability and recognition accuracy compared to state-of-the-art transformer models.
\end{abstract}

\section{Introduction}
Visually grounded Language Acquisition (VLA) is an innate ability of the human brain, which refers to the way children learn their native language, through exploration, observation, and listening (\textit{i.e.,} self-supervision), and without taking language training lessons (\textit{i.e.,} explicit supervision). 
When 2-year-old children repeatedly hear phrases
like ``washing apple'', or ``cutting carrot'', while observing such situations, they quickly learn the meaning of the phrases and their constituent words. More interestingly, they will understand new compositions such as ``washing carrot'' or ``cutting apple'', even before experiencing them. This ability is called compositional generalization~\cite{montague1970, Minsky1988, Lake2017Building}, and is a critical capacity of human cognition. It helps humans use a limited set of known components (vocabulary) to understand and produce unlimited new phrases (\textit{e.g.} verb-noun, adjective-noun, or adverb-verb compositions) that they have never heard before. 
This is also one of the long-term goals of Artificial Intelligence (AI), \textit{e.g.} in robotics, where it enables the robot to understand new instructions by finite verbal communication.



%

Nevertheless, contemporary machine intelligence needs to overcome several major challenges of the task. 
On the one hand, learning compositional generalization without enough language knowledge can be difficult. Existing language models~\cite{lu2019vilbert,chen2019uniter}, are still inadequate at compositional generalization, despite explicit supervision or large-scale language corpora~\cite{marcus1998rethinking,Lake2018,Didac2019}. This mainly caused by their goal is to recognize training examples rather than focusing on what is missing from training data.
On the other hand, the designed model should close the paradigmatic gap~\cite{nikolaus2019} between seen compositions and new compositions. For instance, given seen verb-noun compositions ``1A'' and ``2B'' (the digit indicates verb, the letter indicates noun), 
the model should be able to link seen compositions to new compositions (like ``1B'' or ``2A'') in completely new and unstructured environments.
Few works have focused on addressing the particular challenges of compositional generalization for machine language acquisition~\cite{kottur2016visual,shi2019visually}.
Different from previous work and inspired by the reasoning mechanism in recent deep learning models~\cite{johnson2017inferring,baradel2018object,santoro2017simple}, we address compositional generalization for VLA by incorporating multimodal reasoning into transformer models.

\begin{figure*}[t]
\includegraphics[width=\textwidth]{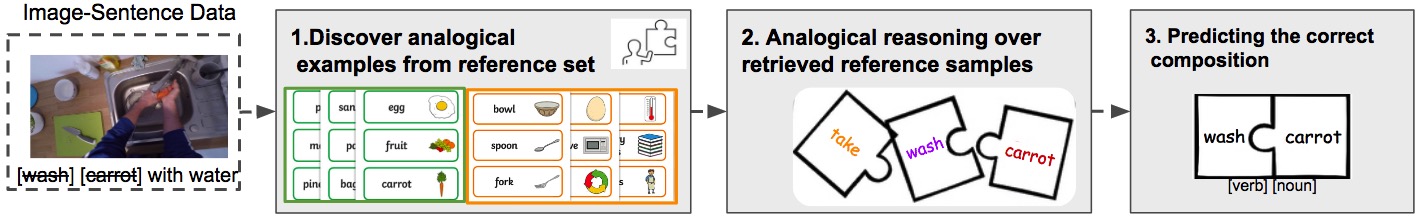}
\caption{We propose a multimodal language acquisition approach inspired by human language learning processes consisting of three steps: association, reasoning, and inference.}
\label{fig:idea} 
\end{figure*}

We bring the power of compositional generalization to state-of-the-art language models by introducing Analogical Reasoning (AR)~\cite{gentner2012analogical,littlemore2008relationship,vosniadou1989similarity}. An analogy is a comparison between similar concepts or situations, and AR is any type of reasoning that relies upon an analogy. The human brain spontaneously engages in AR to make sense of unfamiliar situations in every day life~\cite{vamvakoussi2019}. Inspired by the AR process in the human brain, we design the counterpart for machine language acquisition. To this end, we create a language model that makes sense of novel compositions by recalling similar compositions it has seen during training, and forming analogies and appropriate arithmetic operations to express the new compositions (\textit{e.g.} ``washing carrot'' = ``washing apple' + ``cutting carrot'' - ``cutting apple''). We describe this process in three steps: association, reasoning, and inference, as shown in 
Figure~\ref{fig:idea}.


We take inspiration from masked language modeling such as BERT~\cite{devlin2018bert} to formulate our problem. Given an image (a video frame in our case) and a narrative sentence describing it, we mask a verb-noun composition and ask the model to reconstruct the sentence by predicting those two words. To this end, we propose a novel self-supervised and reasoning-augmented framework, Analogical Reasoning Transformer Networks (\textsc{ARTNet}). \textsc{ARTNet} adopts a multimodal transformer (similar to ViLBERT~\cite{lu2019vilbert}) as its backbone to represent visual-textual data in a common space. Then it builds three novel modules on the top of the backbone that corresponds to the aforementioned AR steps: association, reasoning, and inference.
First, we design Analogical Memory Module (AMM), which discovers analogical exemplars for a given query scenario, from a reference pool of observed samples.
Second, we propose Analogical Reasoning Networks (ARN), which takes the retrieved samples as input, selects candidate \textit{analogy pairs} from the relevant reference samples, and learns proper reasoning operations over the selected analogy pairs, resulting in an \textit{analogy context vector}. 
Third, we devise Conditioned Composition Engine (CCE), which combines the analogy context vector with the representations of the query sample to predict the masked words and complete the target sentence with a novel composition.

We show how \textsc{ARTNet} generalizes to new compositions and excels in visually grounded language acquisition by designing several experiments in various metrics: novel composition prediction, assessment of affordance, and sensitivity to data scarcity. The results on the ego-centric video dataset (EPIC-Kitchens) demonstrate the effectiveness of the proposed solution in various aspects: accuracy, capability, robustness, etc. The code for this project is
publicly available at~\url{https://github.com/XX}.

The main contributions of this paper include the following:
\begin{itemize}
\item We call attention to a challenging problem, compositional generalization, in the context of machine language acquisition, which has seldom been studied. 

\item We propose ideas supported by human analogical reasoning: approximating new verb-noun compositions by learned arithmetic operations over relevant compositions seen before.

\item We propose a novel reasoning-augmented architecture for visually grounded language acquisition, which addresses the compositional generalization problem through association and analogical reasoning.

\item We evaluate the proposed model in various aspects, such as composition prediction, affordance test, and robustness against data scarcity. The results show that \textsc{ARTNet} achieves significant performance improvements in terms of zero-shot composition accuracy, over a large-scale video dataset. 
\end{itemize}

\section{\textsc{ARTNet}: Analogical Reasoning Transformer Networks}

\begin{figure*}[t]
\centering
\includegraphics[width=\textwidth]{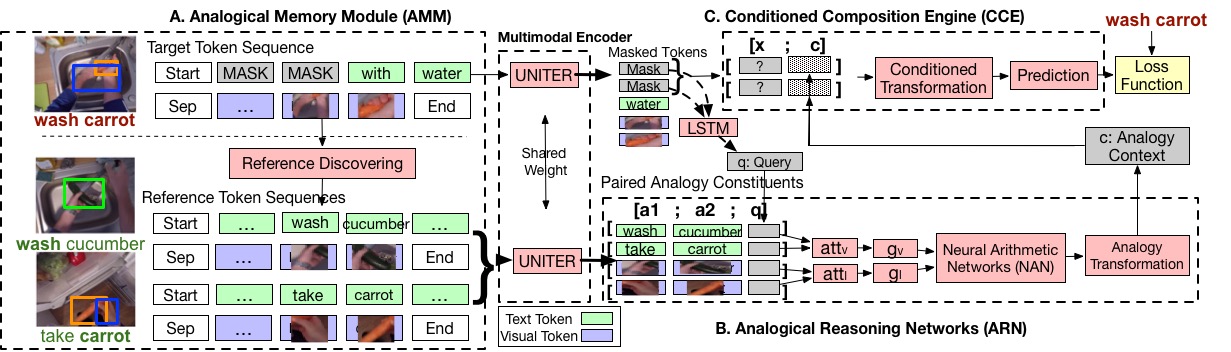}
\caption{\textbf{Analogical Reasoning Transformer Networks (\textsc{ARTNet}):} The proposed reasoning-augmented architecture for compositional generalization via analogical reasoning.}
\label{fig:framework} 
\end{figure*}

Our goal is to develop a framework that can support \textit{compositional generalization} through learning in a visual-text environment. The framework learns to acquire the meaning of phrases and words from image-sentence pairs, and learns to create novel compositions that have never been seen before. We call the proposed framework Analogical Reasoning Transformer Networks (\textsc{ARTNet}), due to its ability to establish analogies with the previously seen, relevant scenarios, and perform reasoning operations to generalize a composition for the new scenario. 
Figure~\ref{fig:framework} illustrates an overview of \textsc{ARTNet}, which is composed of a multimodal encoder backbone, followed by three main modules: Analogical Memory Module (AMM), Analogical Reasoning Networks (ARN), and Conditioned Composition Engine (CCE). 
Given an image and a sentence where the target words are concealed by a special mask symbol, we extract a set of visual and textual features for parsing (\textit{i.e.,} objects and words) and construct token sequences as the input data of a multimodal encoder backbone.
Then AMM takes the pre-computed features of a target sample and compares with a set of reference, to retrieve the most relevant samples. 
Next, ARN attends to the retrieved reference samples and processes them via the proposed reasoning networks, resulting in an analogical context vector.
Finally, CCE reconstructs each masked word by classifying its embedding, conditioned on the analogical context vector.

\subsection{Multimodal Encoder Backbone}
\label{Backbone Model}
The backbone network is responsible for encoding image-sentence pairs into representations in a generic semantic space. 
To achieve this, we utilize the emerging multimodal transformers (\textit{e.g.} UNITER~\cite{chen2019uniter} or ViLBERT~\cite{lu2019vilbert}), which have recently achieved great success in various vision-language tasks. These models pre-process an input image-sentence sample to a set of visual-textual tokens, which are of three types: visual, textual, or special tokens. The visual tokens correspond to regions of interest (RoI) from an object detector, as well as one for the entire image; The textual tokens indicate the words in the given sentence; And the special tokens are used to represent the correspondance of the visual and textual parts, or to conceal (mask) some tokens from the model (in our case verb-noun pairs). We represent visual tokens using a ResNet-18~\cite{he2016deep} pre-trained on ImageNet 2012 dataset~\cite{deng2009imagenet}, and represent the textual or special tokens via one-hot vectors. After collecting the tokens, the model computes an initial embedding for each token, as the sum of a token embedding, a modality embedding (indicating the modality of the token) and a position embedding (indicating the RoI coordinates in image or word position in sentence). 
Next, several layers of transformers~\cite{vaswani2017attention} (attending within tokens of each modality, as well as across modalities) operate over the multimedia tokens, to output a contextualized representation for each token. 
We follow the architecture of UNITER~\cite{chen2019uniter} (since it performs slightly better than ViLBERT and other similar models), but do not use their pre-trained word embeddings or weights. 
Note that since our task is language acquisition, we intentionally do not want to rely on a powerful language model already trained on a large dataset. 
Instead, we train the multimodal encoder from scratch on our data.

\subsection{Analogical Memory Module}
\label{Analogical Reference Discovery}
AMM plays the role of the analogical association. Like finding a useful puzzle piece, we propose AMM to discover the most useful reference samples for the analogical reasoning for a target scenario.
Given a target image-sentence pair (query), where some tokens in the sentence are masked, we randomly select N (N = 200 in our experiments) sample image-sentence pairs from the training data to create a reference pool, and find the Top-K most relevant exemplars from that pool. To this end, we measure a multimodal relevance score between the query and each reference. Here, we use the initial embedding of each token on the query and reference samples as described in the Section~\ref{Backbone Model}. Given a target and a reference sample, we define the multimodal relevance score as a combination of visual and text relevance between the corresponding sets of tokens.
For visual tokens, we compute the mean cosine similarity of every pair of tokens from the query and reference token sets. For the language part, the contextual background words that are not masked can provide linguistic clues for semantic relevance. Thus, we compute the Jaccard Index~\cite{hamers1989similarity} between two sentences as textual relevance.
Specifically, the multimodal relevance score is
\begin{equation}
    s_{vl} = \frac{1}{2} \cdot (\frac{|W_T \cap W_R| }{|W_T \cup W_R|} + \frac{1+\frac{\sum_i{\sum_j{cos(v_{T_{i}}, v_{R_{j}})}}}{N_{v}}}{2})
\label{eq_vlscore}
\end{equation}
where $W_T$ and $W_R$ are the set of target words and reference words, $N_v$ is the number of visual token pairs, and $v_{T_i}$ and $v_{R_j}$ represent the visual embeddings of the $i_{th}$ visual token of the query and the $j_{th}$ visual token of the reference. 
After computing the scores, AMM ranks reference samples with respect to their relevance scores and selects the Top-K most relevant samples for the given query.

\subsection{Analogical Reasoning Networks}
\label{Analogical Reasoning Networks}
Given the retrieved analogical exemplars, we devise a neural network with reasoning ability to enrich the original representation of the masked compositions by making analogies with the seen compositions. 
The idea is to reason the semantic relation mapping between the candidate analogy composition and the target composition. To this end, we represent the masked composition of the target sample as a query vector $q$, by concatenating the multimodal transformer embeddings of the masked words of that composition (typically a verb and a noun from the target sentence) and learning the sequential composition representations based on a Long Short-Term Memory (LSTM)~\cite{zhou2015clstm}.
Next, we apply the multimodal encoder backbone (as mentioned above) on the retrieved analogy samples, and parse each sample into candidate analogy compositions (pairs of tokens).
Since the task is language acquisition, we do not rely on predefined grammar rules or pretrained models to generate the candidate compositions, such as applying part-of-speech tagging and taking each verb-noun pair. 
Instead, we use a heuristic approach to represent visual human-object interactions and linguistic constituents
based on our task domain, which is egocentric cooking videos with short transcripts for each key frame. Specifically, we enumerate all pairs of adjacent words from each retrieved sentence, and all pairs of detected image regions from each retrieved image. 
The resulting set of 
pairs are called analogy pairs hereafter.




The core of ARN are the module networks for analogy attention, analogical reasoning, and analogy transformation. Analogical attention learns the importance of each pair of candidate analogy composition and query vector respectively and generate analogy aggregation from each modality independently.
Analogical reasoning is the networks designed to learn the appropriate arithmetic operations from analogy compositions for reasoning. 
It consists of modality-wise transformations and Neural Arithmetic Logic Units~\cite{trask2018neural}
consists of multiple layers of Neural Accumulator (NAC) ~\cite{trask2018neural}. NAC is a simple but effective operator that supports the ability to learn addition and subtraction. This module is applied on the analogy pairs, and computes a single vector that represents the output of some reasoning operations, optimized for our task through gradient descent. 
Through the analogy transformation, ARN generates the sequential representations of final analogy context vector. 
Specifically, ARN
can be denoted as
\begin{align}
c_a^{m} &= \sum_j \alpha_{ij}^{m} h_{j}^{m},
\alpha_{ij} = \frac{  \exp{a([r_i^{m}; r_{i+1}^{m}; q], [r_j^{m}; r_{j+1}^{m}; q])} }{ \sum_k \exp{a([r_i^{m}; r_{i+1}^{m}; q], [r_k^{m}; r_{k+1}^{m}; q])} } 
&&\text{Analogical Attention,}  \label{eq2} \\
h_c &= f_{NAC}([g_v(c_a^{v}), g_l(c_a^{l})]^T)
&&\text{Analogical Reasoning,}  \label{eq3} \\
c &= LSTM (h_c) 
&&\text{Analogy Transformation,}  
\label{eq4}
\end{align}
where $v$ and $l$ represent the vision and language modalities, $r_i^{m}$ and $r_{i+1}^{m}$ ($m$ is modality indicator) are the image regions or text words of candidate analogical compositions. $g_v$ and $g_l$ are modality transformations that contains two-layer fully connected networks with ReLU activation and 0.5 Dropout, and $T$ represents matrix transpose. 
The output of ARN is the vector $c$, which is called analogical context vector, and will be used to augment the composition representations.

\subsection{Conditioned Composition Engine}
\label{Conditioned Learning Networks}
After analogical reasoning, we create a potentially novel composition based on both the initial comprehension of the given scenario and the result of analogical reasoning. To this end, CCE is designed to enrich the representations of the masked tokens through a conditioned learning network. It takes the analogical context vector as contextual knowledge.
Let $x = \{x_1, ..., m_i, m_{i+1}, ..., x_N \}$ be the input elements of a multimodal encoder, and $m_i^{l}$ or $m_{i+1}^{l}$ are the $l$-th layer of the representations of the masked words. CCE uses the multimodal transformers to transform the embedding features of both each masked word and the analogical context vector. Then it predicts the masked word by aggregating from linguistic clues of all the other unmasked elements. 
The embedding feature $h_i$ of the $i$-th masked word computed by:
\begin{align}
h_i^{l+1} &= W_2^{l+1} \cdot ReLU(W_1^{l+1}[m_i^{l};c] + b_1^{l+1}) + b_2^{l+1} &&\text{, Context-conditioned} \label{eq5}\\
h_i^{l+2} &= GELU(W_i^{l+2}h_i^{l+1} + b_i^{l+2}) &&\text{, Feed-forward} \label{eq6}\\
h_i &= LayerNorm(h_i^{l+2}) &&\text{, Feed-forward} \label{eq7}
\end{align}
where $W_1^{l+1}$, $W_2^{l+1}$, $b_1^{l+1}$ and $b_2^{l+1}$ are learnable weights and biases for the context-conditioned transformation. $W_i^{l+2}$ and $b_i^{l+2}$ are learnable weight and bias for feed-forward transformation, respectively. Given the contextual representation of the masked word $h_i$, the model predicts the masked word by multiplying its contextual representation with a word embedding matrix $\phi$ which is trained with the rest of the network, $\hat{w_i} = \phi_w^T h_i$.




\subsection{Learning Objectives}
\label{Learning Objectives}
\textbf{Masked Composition Acquisition:} 
Our model learns to perform language acquisition by filling in masked words. At both train and test time, we give the model a target example, with a reference set sampled from the training dataset. 
It should be pointed out that the mask policy is different at training time and test time (details in Section~\ref{sec:exp}). During training time (including validation), we randomly mask multiple words and visual tokens. However, during test time, we only mask one verb-noun composition.

\textbf{Objective Function:} 
Here we define the objective function. We train the model to acquire words by directly predicting them. We measure the Cross-Entropy loss between the predicted word $\hat{w_i}$ and true word $w_{i}$ over a vocabulary of size $C$, denoted as $\mathcal{L}_{\textrm{l}}$. This objective is the same as in the original BERT \cite{devlin2018bert}. We also learn visual reconstruction via a regression task. The visual loss $\mathcal{L}_{\textrm{v}}$ is a Triplet Metric loss~\cite{weinberger2006distance} to force a linear projection of $v_{i}$ to be closer to $\phi_v(v_i)$ than $\phi_v(v_{k \neq i})$, and $\phi_v(\cdot)$ is the visual representation networks ResNet.
Because the objectives are complementary, we train the parameters of the proposed model by minimizing the sum of losses:
\begin{align}
    \min_\Theta \; \left( -\sum_{i}^{C}w_{i} \log (\hat{w_i}) + \lambda \max(0,m + \left \| v_{i} - \phi_v(v_i) \right \| - \left \| v_{i} - \phi_v(v_{k \neq i}) \right \|) \right)
\end{align}
where $\lambda \in \mathbb{R}$ is the parameter to balance the loss terms (modalities), $m$ is the triplet margin and $\Theta$ represents the trainable parameters of the entire network.

\section{Experiments}
We compare our method (\textsc{ARTNet}) and baselines on new and seen composition acquisition tasks. To demonstrate the 
quantitative and qualitative results, we evaluate our method in a variety of aspects including performance comparison, affordance test, incremental data setting, and case study.

\subsection{Experiment Settings}
\label{sec:exp}
\paragraph{Dataset}
We use the instruction video dataset EPIC-Kitchens~\cite{Damen2018EPIC} in our experiments. The dataset consists of $55$ hours of egocentric videos across $32$ kitchens.
Each video clip has a narrative sentence, with a vocabulary that includes but is not limited to $314$ unique verbs and $678$ unique nouns. 
We create pairs of images and sentences by selecting the key frames of each video clip and their corresponding sentences.
For each video frame, we use the object bounding boxes officially provided with EPIC-Kitchens, which are produced by running a Faster R-CNN. We discard the object labels due to strict constraints in the language acquisition scenario. There are about 3-6 objects per image.
Since we attempt to analyze compositional generalization, we create a train-test split with $140K$/$30K$ training/testing instances where the test set has both seen and new compositions. We partition the dataset to ensure new compositions used in testing have never been seen in training.

\paragraph{Evaluation Metrics}
To compare the performance of our model against baselines, we evaluate the ability to acquire new or seen compositions (\textit{e.g.} verb-noun word pairs).
During testing, the model takes paired image-sentence data as input, and the target verb-noun composition is replaced with a special ``mask'' token. 
The objective is to predict the masked words from the complete vocabulary of all words (unaware of the part of speech). 
We adopt Top-1 and Top-5 accuracy to measure the performance, which calculates the average accuracy over the predicted nouns and verbs with the top N highest probability. 
The prediction is correct when both noun and verb are correctly predicted.


\paragraph{Implementation Details}
In all experiments, training with batch-size 256 is conducted on 2 GPUs for 200 epochs. 
 The transformer encoder in all models has the same configuration: 4 layers, a hidden size of 384, and 4 self-attention heads in each layer. 
Besides, we use AdamW optimizer ~\cite{loshchilov2017decoupled} with the learning rate 3e-5. We set the update coefficients, for averages of gradient and its square ($\beta_1$, $\beta_2$), and $\epsilon$ on denominator as 0.9, 0.999, 1e-4.
During training, we mask out text/image tokens $\frac{1}{3}$/$\frac{1}{6}$ of the time and follow the same setting of random masking strategy with BERT ~\cite{devlin2018bert}. 
During training and testing, for each target sample, we randomly select 200 remaining samples as the corresponding reference set and utilize Top-K (K = 3) reference samples to get involved in analogical reasoning.


\begin{table*}[t]
\centering
\begin{tabular}{c|c|c|c|c}
\hline
\multirow{2}{*}{\textbf{Models / Metrics} (\%)}  & \multicolumn{2}{c}{\textbf{New Composition}} & \multicolumn{2}{|c}{\textbf{Seen Composition}} \\ \cline{2-5}
          & \textbf{Top-1 Acc.}       & \textbf{Top-5 Acc.}        & \textbf{Top-1 Acc.}        & \textbf{Top-5 Acc.}        \\ \hline
BERT (from scratch)~\cite{devlin2018bert}           &     2.25      & 6.85             &  6.47            & 26.42             \\ 
BERT (pre-trained)~\cite{devlin2018bert, zhu2015aligning, chelba2013one}               &    2.06          & 10.41        &  7.89           & 27.10             \\ 
Multimodal BERT (from scratch)~\cite{chen2019uniter} & 5.97             & 36.63             & \textbf{23.22}             & \textbf{57.81}             \\ 
\textbf{Proposed \textsc{ARTNet}} (from scratch)   & \textbf{7.10 (+1.13)}     & \textbf{40.90 (+4.27)}      & 22.45             & 57.45             \\ \hline          
\end{tabular}
\caption{\textbf{Performance Comparison:} Top-1 and Top-5 accuracy comparison on new and seen composition acquisition tasks.}
\label{performance} 
\end{table*}

\subsection{Performance Comparison with Baselines}
We compare our model with state-of-the-art methods of language modeling or multimodal language modeling. We consider three variants of the BERT family ~\cite{devlin2018bert, chen2019uniter}:

\textbf{BERT (from scratch or pre-trained)} is a powerful language-only transformer model~\cite{devlin2018bert} that achieved state-of-the-art performance in multiple language learning tasks. 
We adopt two learning settings: (1) train the BERT model from scratch on the sentences of task data; (2) utilize the pre-trained BERT model~\cite{zhu2015aligning, chelba2013one} and fine-tune BERT on the task data. 
Note that the pretrained BERT has the advantage of large-scale data, which is not realistic in language acquisition scenario.

\textbf{Multimodal BERT (from scratch)} is a multimodal transformers model by using both visual and textual data (paired image-sentence data) as inputs. It has the same architecture with our backbone UNITER~\cite{chen2019uniter}, and we adopt a pre-trained ResNet-18~\cite{he2016deep} to extract visual features. 
The model is trained on our task data from scratch (without using the pre-trained model), because we focus on the setting of learning from scratch in VLA task. 

\begin{figure}[ht]
\centering
\begin{minipage}{0.58\linewidth}
\centering
\includegraphics[width=\textwidth]{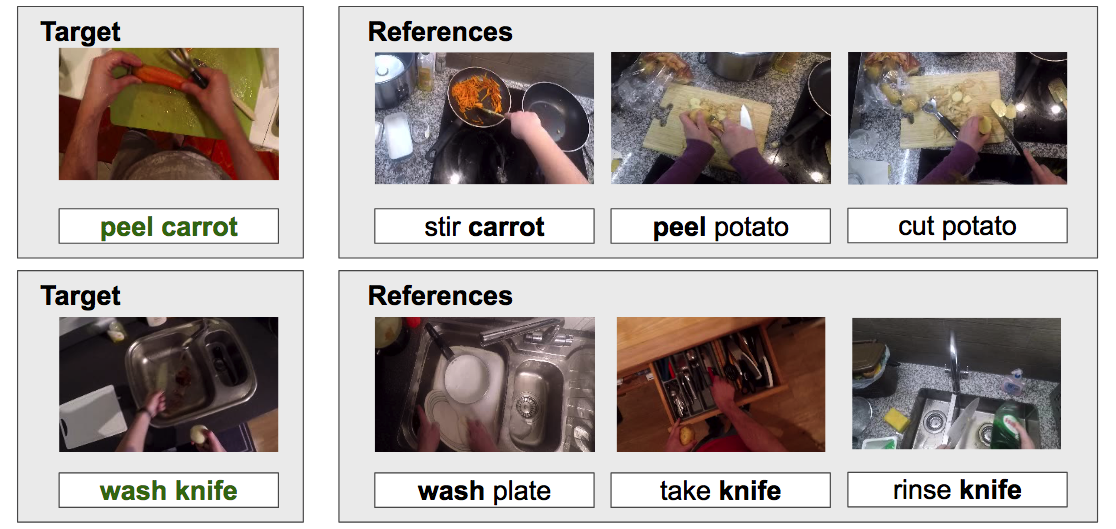}
\caption{The target sample and top samples discovered from the reference set.}
\label{fig:tar_ref_vis}
\end{minipage}
\begin{minipage}{0.41\linewidth}
\centering
\includegraphics[width=\textwidth]{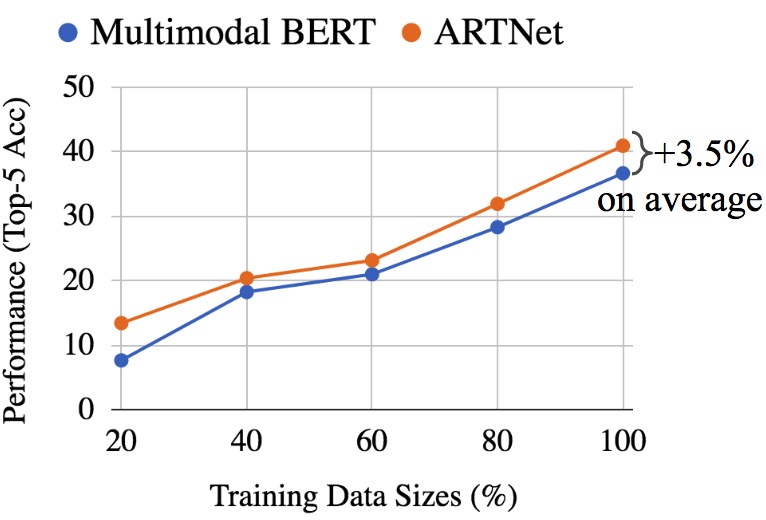}
\caption{\textbf{Robustness:} Performance curve with different sizes of training data.}
\label{fig:per_diff_size}
\end{minipage}
\end{figure}



Quantitative and qualitative results are illustrated in Table~\ref{performance} and Figure~\ref{fig:tar_ref_vis}, respectively. For new composition prediction, as shown in Table~\ref{performance}, our model can achieve $1.13\%$ and $4.27\%$ improvement over state-of-the-art baseline on Top-1 and Top-5 accuracy respectively. For seen composition prediction, the proposed model is nearly unchanged. Such improvement results from the analogical reasoning based on selected proper reference samples.
As shown in Figure~\ref{fig:tar_ref_vis}, the model can successfully retrieve relevant reference samples that which contain analogical pairs.

\subsection{Experiments on Different Training Sizes}
To evaluate our model on low-data regime and simulate the language acquisition process of a child, we train on different scales of data. Specifically, we consider 5 different training data size percentages (TPs) (100\%, 80\%, 60\%, 40\% and 20\%), and plot the performance of our method compared to the stringest baseline in Figure~\ref{fig:per_diff_size}. 
Our \textsc{ARTNet} can achieve 13.44\% Top-5 accuracy with only 20\% of training data, which has larger gap with the baseline, suggesting our stronger generalization ability.



\subsection{Affordance Test Evaluation}
We propose a novel way to evaluate the model for physical capacity, called ``Affordance Test''. This evaluation aims at assessing whether the predicted results follow human commonsense in the real world. 
Affordance is a binary relationship between each noun and verb, which indicates whether a certain verb is typical to involve a certain noun as an argument~\cite{baber2018designing, gibson2014ecological}. 
For examples, ``cutting pizza'' and ``cutting carrot'' happen in our daily life, but we never see ``cutting plate'' or ``cutting water''.
To demonstrate our method has acquired the language priors of affordance, we annotate the Top-1 verb-noun compositions predicted by each model given any test image. 
The type-mismatched compositions, such as ``meat apple'' (noun-noun) or ``peel cut'' (verb-verb), are defined as unaffordable compositions.
We annotated 8400 compositions in total, 5704 of which appear in our dataset. 


\begin{table*}[t]
\parbox{.4\textwidth}{
\begin{tabular}{c|ccc}
\hline
Composition Acquisition & \multicolumn{3}{c}{Affordance Accuracy} \\ 
\hline
Methods/Settings        & Overall    & New Comp    & Seen Comp    \\
\hline
BERT (w/o vision)~\cite{devlin2018bert}    & 81.67\%    & 81.59\%     & 81.72\%      \\
Multimodal BERT~\cite{chen2019uniter}       & 85.79\%    & 84.78\%     & 86.36\%      \\
\textbf{Proposed \textsc{ARTNet}}    & \textbf{86.48}\%    & \textbf{86.01}\%     & \textbf{86.75}\%      \\
\hline
\end{tabular}
}
\hspace{\fill}
\parbox{.3\textwidth}{
\caption{\textbf{Affordance Test:} Performance comparison of affordance accuracy.}
\label{comp_affordance} 
}
\end{table*}

For the affordance test comparison, we show the Top-1 affordance accuracy (the ratio of the number of affordable predictions to all) for each model in Table \ref{comp_affordance}. 
By comparing the first row and the second row, we find that the overall affordance accuracy increases by $4.12\%$ if the model takes advantage of visual information. Visual clues serve as a guide for the model to learn better about physical properties of objects, and hence their affordance. 
Moreover, \textsc{ARTNet}, empowered by analogical reasoning, outperforms all baselines significantly.

\begin{figure*}[ht]
\parbox{.6\textwidth}{
\includegraphics[width=0.62\textwidth]{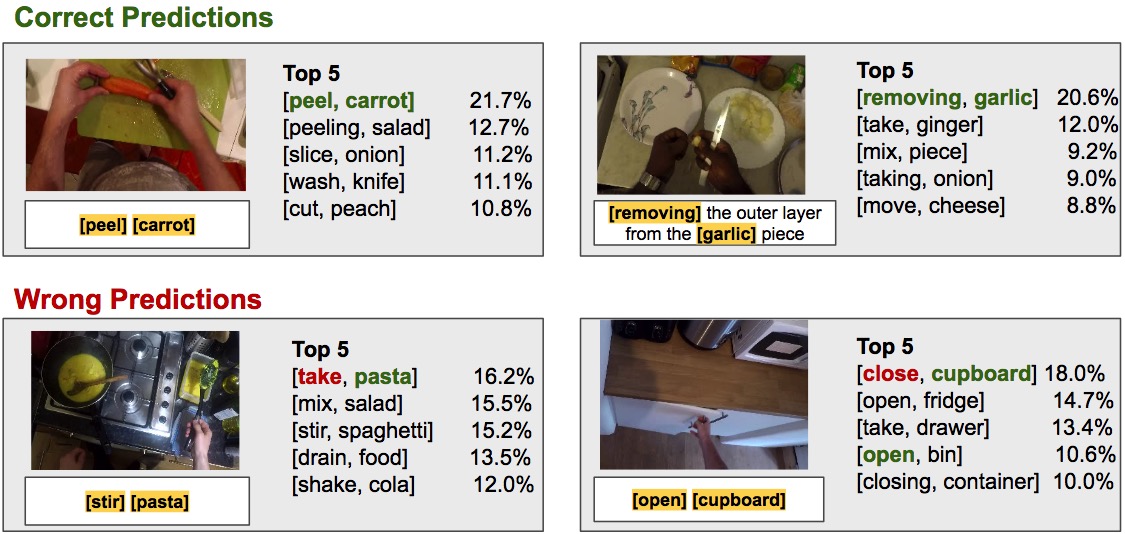}
}
\hspace{\fill}
\parbox{.37\textwidth}{
\caption{\textbf{Case Study:} Top-5 prediction results for new compositions (highlight words are masked ground-truth compositions to be predicted).\label{fig:topk_vis} }
}
\end{figure*}

\subsection{Case Study}
\begin{figure*}[ht]
\centering
\subfigure{
\label{Aff.sub.1}
\includegraphics[width=0.4\textwidth]{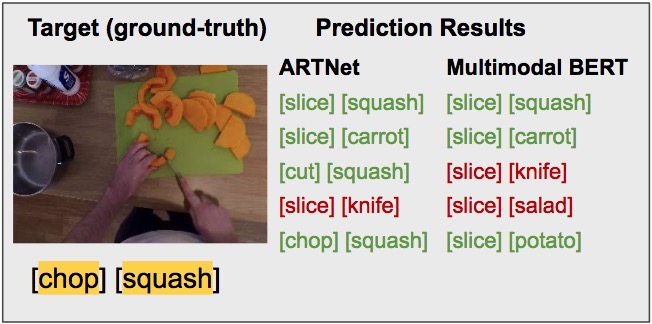}}
\subfigure{
\label{Aff.sub.2}
\includegraphics[width=0.58\textwidth]{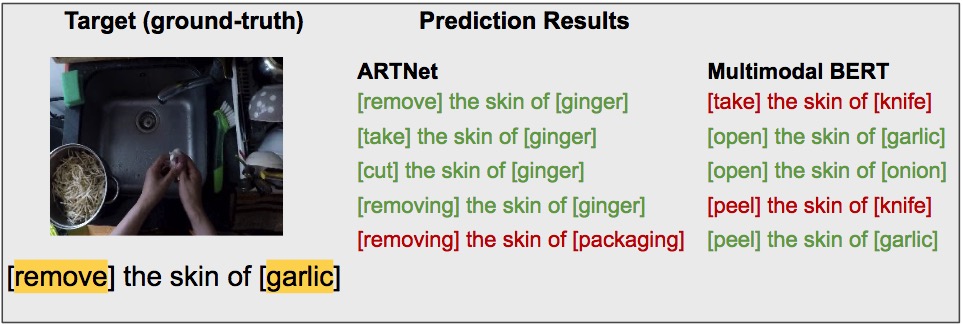}}
\caption{\textbf{Case Study:} The affordance test results of two examples. The two words of each example in brackets are masked and to be predicted as the target composition. We show the Top-5 prediction results for each example. Green or red color indicates affordable or unaffordable prediction.}
\label{affordance_example}
\end{figure*}

To analyze the strengths and failure cases, we select some examples shown in Figure~\ref{fig:topk_vis}.
Intuitively, vague visual clues such as ambiguous action and tiny objects, are hard to recognize. However, we observe \textsc{ARTNet} has a unique ability to disambiguate such difficult cases, as shown in the top row of Figure~\ref{fig:topk_vis}.
Nevertheless, the model fails in some cases, such as take/stir and close/open. This is mainly due to the fact that we used still keyframes rather than dynamic videos, which discards motion information. Future work should incorporate spatio-temporal features to alleviate this limitation.
Moreover, we provide several cases with Top-5 prediction results to show the affordance and rationality of our method's prediction. 
Considering Figure \ref{affordance_example}, we observe that not only the Top-1 results, but also most of the Top-5 are in line with human commonsense, and superior to the baseline. 

\section{Related Work}
\paragraph{Compositional Generalization}
Compositional generalization is an important open reasoning challenge~\cite{keysers2019measuring, chang2018automatically, loula2018rearranging} and a priority to achieve human-like abilities~\cite{bastings2018}, regarding the ability to compose simple constituents into complex structures~\cite{nikolaus2019}.
From the traditional view, numerous studies utilize linguistic principle to explore compositionality~\cite{mitchell2008vector, baroni2010nouns}.
With recent advances, there are more attempts to solve compositional generalization with neural networks for the tasks with synthetic commends or interactions~\cite{lake2019compositional, russin2019compositional, li2019compositional, kato2018compositional}.  
The current machine intelligence still lacks compositional generalization ability, because they are prone to fail on such tests in realistic or natural scenarios~\cite{nikolaus2019,loula2018rearranging,keysers2019measuring}. Moreover, several recent works also try to enhance compositional generalization for natural language~\cite{nikolaus2019} with pretrained representations or lexical knowledge. 
But there are few works addressing this emerging and valuable challenge in language acquisition. 

\paragraph{Visually Grounded Language Acquisition}
Visually grounded Language Acquisition (VLA) is a task of acquiring concrete language constituents from scratch within a visual-text environment. 
Although the works of grounded language learning achieve the good progress in many tasks (such as visual captioning~\cite{yu2013grounded,kiela2017learning,ross2018grounding}, visual navigation\cite{anderson2018vision} or robotics~\cite{matuszek2018grounded}), they try to utilize explicit semantic knowledge or pretrained word representations on large-scale of corpus data~\cite{lu2019vilbert,chen2019uniter}.
Several recent works further address acquiring concrete language knowledge including word representation~\cite{kottur2016visual,suris2019learning}, compositional semantics~\cite{jin2020visually} or sentence syntax~\cite{shi2019visually} from scratch via visual clues, which are more related to our task. 
Our work seeks to improve, in the form of reasoning, the generalization to new compositions.

\paragraph{Reasoning-Augmented Neural Networks}
Arming neural networks models with the reasoning ability enables them to learn the relation or structure behind them. 
Reasoning-augmented models add components to neural network models to facilitate the reasoning processes in such models~\cite{johnson2017inferring}. 
One form of reasoning-augmented models is module networks~\cite{andreas2016learning,andreas2016neural}. They use a parser~\cite{andreas2016learning,andreas2016neural,hu2017modeling,hu2017learning} or a pretrained program generator~\cite{johnson2017inferring,yi2018neural} of a question to determine the architecture of the network. The final network with the trained modules that can execute the generated programs. But the reasoning ability strongly depends on the available program supervisions, which is easy to be a bottleneck of the entire system. 
Another form of the reasoning-augmented model is plug-in reasoning networks, which have been applied for several applications from visual reasoning~\cite{santoro2017simple,baradel2018object,hudson2018compositional} to reasoning about physical systems~\cite{battaglia2016interaction}. These methods are better characterized by memorization rather than by systematic abstraction or generalization~\cite{trask2018neural,russin2019compositional}. 
In our work, the proposed reasoning networks is a plug-in reasoning module and does not require reasoning-specific supervisions.


\section{Conclusion}
In this paper, we take a step towards visually grounded language acquisition, by studying the problem of compositional generalization in the state-of-the-art multimedia language models. Inspired by the human brain's analogical reasoning process, we propose to form new compositions by recalling observed compositions and relating them to the new scenario through learned arithmetic operations. Our proposed reasoning-augmented method, Analogical Reasoning Transformer Networks, achieves superior compositional generalization capability compared to the state-of-the-art transformers, which results in significant and stable performance gains in unseen compositions over a large-scale instructional video dataset.

\section{Broader Impact}
Our work is inspired by the human process of language acquisition, thus our approach can help demonstrate the merit of machine learning grounded to the human-learning processes, as opposed to the alternative focused more on data-driven learning. It helps motivate more research leveraging knowledge across disciplines, in this case, connecting cognitive science and machine learning. If successful, the system can help develop robotics communication and intelligent information triage system in low-resource language settings. Like other AI systems, the potential negative impact will be loss of human job opportunities displaced by the automatic agents. The consequence caused by  systems failures could be the biased predictions, due to bias introduced in the training data. We have evaluated certain aspects of these effects in the paper (\textit{e.g.}, affordance properties of the predictions and performance sensitivity to data scarcity), and will further explore other dimensions in the future.

\section{Acknowledgement}
This work was supported by the U.S. DARPA GAILA Program No.HR00111990058. The views and conclusions contained in this document are those of the authors and should not be interpreted as representing the official policies, either expressed or implied, of the U.S. Government. The U.S. Government is authorized to reproduce and distribute reprints for Government purposes notwithstanding any copyright notation here on.
\bibliographystyle{ieee_fullname}
\bibliography{bobby}

\end{document}